\pdfoutput=1
\documentclass[submission,copyright,creativecommons]{eptcs}
\providecommand{\event}{ICLP 2024} 

\usepackage{amssymb}
\usepackage{hyperref}
\usepackage{booktabs}
\usepackage[english]{babel}

\usepackage{epigraph}

\usepackage{graphicx}
\usepackage{mathptmx}
\usepackage{tikz-qtree}
\usepackage{filecontents}
\usepackage{csvsimple}
\usepackage{amsfonts,amsmath}
\usepackage{subfig} 
\usepackage{url}
\usepackage{verbatim}
\usepackage{color}
\usepackage{amsmath} 
\usepackage{proof}

\renewcommand{\phi}{\varphi}

\input prolog.tex

\definecolor{lgray}{gray}{0.95}
\definecolor{lblue}{rgb}{0.90,0.90,1.00}
\definecolor{lyellow}{rgb}{1.00,1.00,0.70}

\usepackage{listings}
\newtheorem{prop}{Proposition}
\newtheorem{ex}{Example}

\lstnewenvironment{codex}
    {\lstset{}%
      \csname lst@SetFirstLabel\endcsname}
    {\csname lst@SaveFirstLabel\endcsname}
\newcommand{\BI}[0]{\begin{itemize}}
\newcommand{\EI}[0]{\end{itemize}}
\newcommand{\I}[0]{\item}
\newcommand{\BE}[0]{\begin{enumerate}}
\newcommand{\EE}[0]{\end{enumerate}}

\newcommand{\BX}[0]{\begin{ex}}
\newcommand{\EX}[0]{\end{ex}}
\newcommand{\BV}[0]{\begin{verbatim}}
\newcommand{\EV}[0]{\end{verbatim}}

\newcommand{\BP}[0]{\begin{prop}}
\newcommand{\EP}[0]{\end{prop}}

\newcommand{\BEQ}{\begin{equation}}
\newcommand{\EEQ}{\end{equation}}

\newcommand{\FI}{\rightarrow}

\newcommand{\OR}{\vee}
\newcommand{\NOT}{\neg}

\newcommand{\BC}[0]{\begin{center}}
\newcommand{\EC}[0]{\end{center}}

\newcommand{\BF}[0]{\begin{filecontents*}{data.csv}}

\newcommand{\BQ}[0]{\color{blue}\begin{quote}}
\newcommand{\EQ}[0]{\end{quote}\color{black}}

\def \bscale1 {0.25}
\def \bscale {0.25}

\newcommand{\FIG}[4]{
\begin{figure}[htbp]
\centering
{\includegraphics[scale=#3]{#4}}
\caption{#2}
\label{#1}
\end{figure}
}

\begin{document}
\title{
On LLM-generated Logic Programs and their Inference Execution Methods
}

\author{{\bf Paul Tarau}$^{0000-0001-7192-9421}$
\institute{University of North Texas
}
\email{paul.tarau@unt.edu}
}

\def \authorrunning{Paul Tarau}
\def\titlerunning{On LLM-generated Logic Programs and their Inference Execution Methods}

\date{}
\maketitle

\begin{abstract}
Large Language Models (LLMs) trained on petabytes of data are highly compressed repositories of a significant proportion of the knowledge accumulated and distilled so far.
In this paper we study techniques to elicit this knowledge in the form of several classes of logic programs, including propositional Horn clauses, Dual Horn clauses, relational triplets and Definite Clause Grammars.
Exposing this knowledge as logic programs enables sound reasoning methods that can verify alignment of LLM outputs to their intended uses and extend their inference capabilities. 
We study new execution methods for the generated programs, including soft-unification of abducible facts against LLM-generated content stored in a vector database as well as GPU-based acceleration of minimal model computation that supports  inference with large LLM-generated programs.

\end{abstract}

\section{Introduction}

While the multi-step dialog model initiated by ChatGPT is now available from  a few dozen online or locally run open source and closed source  LLMs,
it does not cover the need  to efficiently extract salient information from an LLMs ``parameter-memory'' that encapsulates in a heavily 
compressed form the result of training the model on trillions of documents and  multimodal data. 

Steps in this direction have been taken, relying on ground-truth involving additional  information sources (e.g., collections of reference documents or use of web search queries). Among them, we mention work on  improving performance of Retrieval Augmented Generation (RAG) systems \cite{rag20} by recursively embedding, clustering, and summarizing chunks of text for better hierarchical LLM-assisted summarization \cite{raptor24}, multi-agent hybrid LLM and local computation aggregators \cite{langchain} and deductive verifiers of chain of thoughts reasoning \cite{ling2023deductive}.

A more direct approach is recursion on LLM queries, by chaining 
the LLM's distilled output as input to a next step
 and casting its content and interrelations in the form of logic programs,
to automate and focus this information extraction with minimal human input \cite{tarau2023automation,flops24}.
Like in the case of typical RAG architectures \cite{rag20,raptor24}, this process can rely on external ground truth but it can also use new LLM client instances as ``oracles'' deciding the validity of the synthesized rules or facts.

With focus on automation of this unmediated salient knowledge extraction from the LLM's parameter memory and its encapsulation in the form of synthesized logic programming code, we will extend in this paper the work initiated in \cite{tarau2023automation,flops24} with:
\BI
\I new LLM input-output chaining mechanisms
\I new types of generated logic programs
\I new relational representations elicited from LLM output steps
\I scalable execution mechanisms that accommodate very large logic programs at deeper recursion levels
\I soft-unification based execution of LLM-generated logic programs as a principled encapsulation of the RAG retrieval process
\EI
The rest of the paper is organized as follows.
Section \ref{rec} overviews the DeepLLM architecture described in \cite{flops24} and its new extensions supporting the results in this paper.
Section \ref{horn} overviews the generation of Horn clause programs with the online DeepLLM app.
Section \ref{dual} explains the LLM-based generation of Dual Horn clause programs and their uses to explore counterfactual consequences and theory falsification,
Section \ref{grams} introduces the use of DCG grammars as a representation of LLM-generated answer and follow-up question pairs.
Section \ref{torch} describes fixpoint and GPU-supported minimal model computation for the generated programs.
Section \ref{rels} describes relation-extraction and visualization from the minimal models of the LLM-generated propositional programs.
Section \ref{soft} introduces the soft-unification based encapsulation of the information retrieval against facts extracted from authoritative document collections.
Section \ref{rel} discusses related work and Section \ref{conc} concludes the paper.

\section{Recursive exploration of LLM dialog threads}\label{rec}

Generative AI, with often human-like language skills is shifting focus from typical search engines to more conversational interactions. Yet, the challenge remains that humans must still process and verify this information, an often tedious task. 

Our answer to this, as implemented in the DeepLLM system
is to automate the entire process. We start with a simple ``initiator goal'' and let the LLM dive recursively in its parametric memory and deliver a detailed answer focused on the initiator and the trace of this chain of steps summarized as the short term-memory maintained via its API. This automation also helps to minimize common issues like inaccuracies, made-up information, and biases that are often associated with LLMs.

We refer to \cite{flops24,tarau2023automation} for details of implementation of the DeepLLM 
system, as well as to its open-source 
code\footnote{\url{https://github.com/ptarau/recursors/}} 
and its online demo\footnote{\url{https://deepllm.streamlit.app}}.

The DeepLLM system's active components (subclasses of the Agent class) are Interactors, Recursors, and Refiners:
\BI
\I Interactors manage input prompts and task breakdown
\I Recursors handle iterative exploration of subtasks
\I Refiners enhance clarity and relevance of LLM responses
\EI

To validate its reasoning steps, the system also relies on stored knowledge resources: 
\BI 
\I Ground truth facts: sentences collected from online sources or local documents
\I Vector store: enabler of ``semantic search'' via embeddings of sentences
\EI

Starting from a succinct prompt (typically a nominal phrase or a short sentence describing the task) an Interactor will call the LLM via its API, driven by a Recursor that analyzes the LLM's responses and activates new LLM queries  as it proceeds to refine the information received up to a given depth. 

Refiners are Recursor subclasses that rely on semantic search in an embeddings store containing ground-truth facts as well as on oracles implemented as specialized Interactors that ask the LLM for advice on deciding the truth of, or the rating of hypotheses. Besides returning a stream of answers, Recursors and Refiners compile their reasoning steps to a propositional Horn clause program available for inspection by the user or subject for execution and analysis with logic programming tools (in particular, with our model builder -- a fast propositional Horn clause theorem prover).

\section{Generating propositional Horn clause programs with the DeepLLM app}\label{horn}

\FIG{deepllm}{DeepLLM app}{0.60}{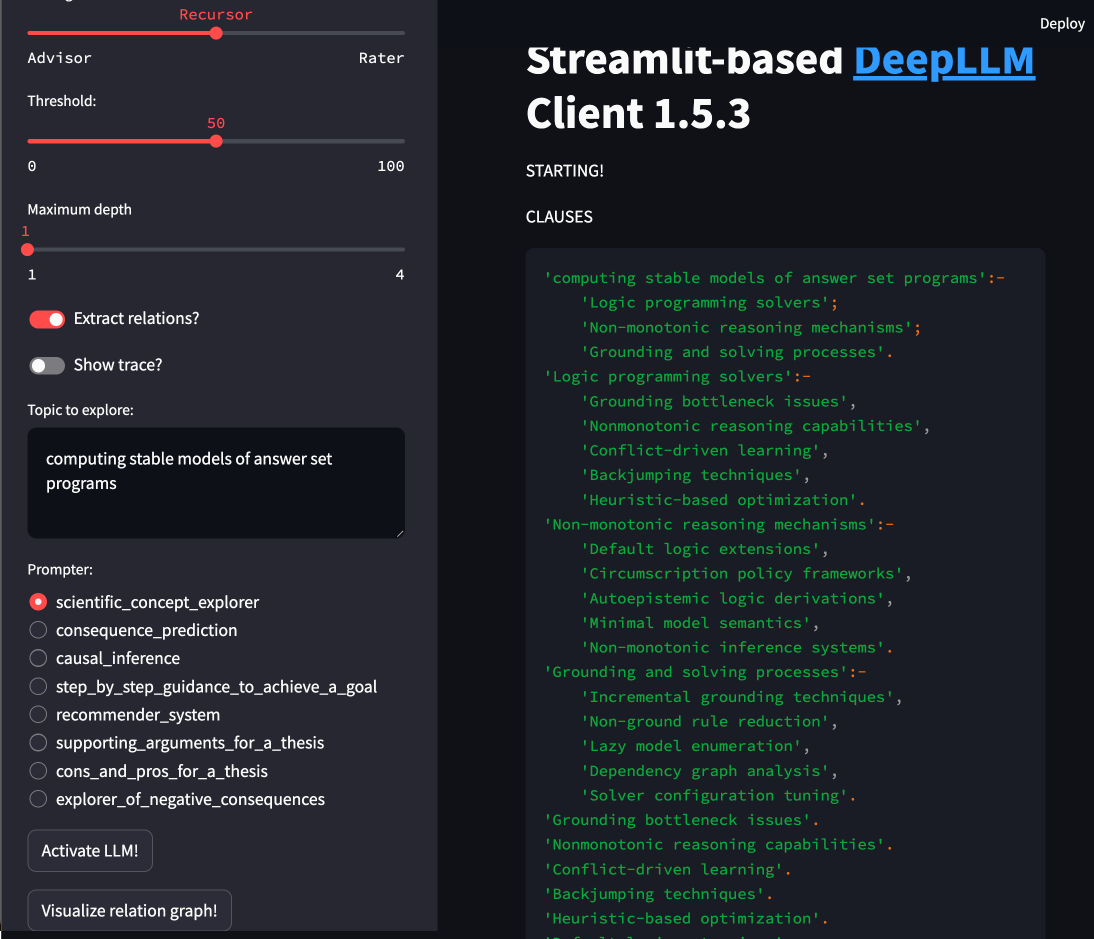}

We refer to \cite{tarau2023automation} for an extensive list of LLM-generated Horn clause programs. We will just briefly describe here the DeeLLM app (see {\bf Fig. \ref{deepllm}}) that we will use for generating our logic programs.
In the case of the interaction shown in {\bf Fig. \ref{deepllm}}, the initiator goal
``computing stable models of answer set programs'' starts the ``scientific concept explorer'' option and generates in the right side window a Horn clause program describing successive refinements of the initiator goal.

The DeeLLM app is written with the {\tt Streamlit}\footnote{\url{https://streamlit.io/}} webapp generator and offers the choice between GPT4, GPT3.5 or a local LLM, running as a server and supporting an OpenAI compatible API. It then lets the user choose between the Recursor, Advisor and Rater agents, providing for the latter a threshold level slider. The threshold informs the Rater oracle to accept or reject a generated rule head or fact (the higher the threshold the stricter the accept decision). Options to set the maximum recursion depth and activate relation extraction and visualization are also available.

The application starts once the user enters the topic to explore, chooses the prompter template and activates the LLM. Besides the output produced in the right window, when run locally, it saves the generated logic program and its computed minimal model as Prolog code files.

\section{Generating propositional Dual Horn clause programs}\label{dual}

A {\em Dual Horn clause} is a disjunction of literals with at most one negative literal (or exactly one if it is a definite Dual Horn clause).
A {\em Dual Horn clause Program}  is a conjunction of Dual Horn clauses.
We represent a Dual Horn clause like
$
\NOT p_0 \OR  p_1 \OR \ldots \OR   p_n
$
in an equivalent implicational form
$
p_0 \FI p_1 \OR  \ldots \OR p_n
$,
similarly to Prolog's representation of Horn clauses.
We adopt a
Prolog-like syntax, with $\rightarrow$ represented as ``\verb~=>~'' and $\OR$ represented as ``;''.
Note also that ``\verb~s => false~'' represents a negated fact the same way as ``\verb~s :- true~'' would represent a positively stated fact.

The objective of Dual Horn programs is to describe (constructively) why something is not true  i.e., to falsify the initiator goal
by back-propagating from its negative (or more generally, undesirable, unwanted, harmful, impossible, etc.,) consequences.

For instance, from a clause like \verb~p => q ; r ; s~, assuming that {\tt p} were true, we would infer that at least one of {\tt q} , {\tt r} and {\tt s} should be true. The contrapositive is that if  {\tt q} , {\tt r} and {\tt s} are all false, then {\tt p} should be false as well.
Like in the case of SLD-resolution on Horn clauses, this triggers a goal oriented process where successful falsification of all consequences results in falsification of the ``counterfactual'' hypothesis that initiated the process.

By instructing the LLM to infer the negative consequences of the DeepLLM initiator goal, we can obtain a Dual Horn program.
\BX  The Dual Horn clauses (recursion level = 0) with heads (starting with  consequences of {\tt `tailgate when driving'}) in the clause body are:

\begin{code}
'tailgate when driving' =>
    'Increased accident risk';
    'Reduced reaction time'.
'Increased accident risk' =>
    'Higher insurance premiums';
    'Severe injury likelihood';
    'Vehicle damage costs';
    'Legal consequences';
    'Emotional trauma impact'.
 \end{code}

 \begin{code}   
'Reduced reaction time' =>
    'Increased accident risk';
    'Delayed braking response';
    'Higher collision likelihood';
    'Compromised driving safety'.
\end{code}
The negative facts (unexplored recursion level = 1 goals) are:
\begin{code}
'Higher insurance premiums' => false.
'Severe injury likelihood' => false.
'Vehicle damage costs' => false.
'Legal consequences' => false.
'Emotional trauma impact '=> false.
\end{code}

 \begin{code} 
'Delayed braking response' => false.
'Higher collision likelihood' => false.
'Compromised driving safety' => false.
\end{code}
Note that ``\verb~=> false~'' marks things that we do not want to happen, from where the same backpropagates to the initiator {\tt `tailgate when driving'}.
Our compilation algorithm\footnote{\url{https://github.com/ptarau/TypesAndProofs/blob/master/symlp/compile_clauses.pro}} will transform this into a definite program placed in module {\tt false} that can be queried with:
\begin{codex}
?- false:'tailgate when driving'.
true
\end{codex}
\EX
Its successful falsification could then advise a car driving program or person to avoid the aforementioned  behavior.

A more interesting exploration (at recursion level=2) of negative consequences in the form of a Dual Horn clause program\footnote{full code at \url{https://github.com/ptarau/output_samples/tree/main/deepllm}}
reveals persuasive counter-arguments to unwise political decisions.
\BX
A few unwanted consequences at descent level 2 for `{\tt loosing the FED\_s independence}':
\begin{code}
'loosing the FED_s independence' => 
    'Increased political influence on monetary policy'. 
...
'Increased political influence on monetary policy' =>
    'Politicized interest rates';
    'Short-term economic manipulation';
    'Eroded investor confidence'; 
    'Heightened market volatility';
    'Policy-driven inflation risks'.
 ...
 'Eroded investor confidence' => 
    'Market volatility';
    'Capital flight';
    'Reduced foreign investment'.
\end{code}
\EX

\section{From Self-generated follow-up question-answer chains to DCG grammars}\label{grams}

DeepQA\footnote{\url{https://github.com/ptarau/recursors/tree/main/deepQA}} 
(see {\bf Fig. \ref{consneg}}) is a DeepLLM-based application that explores recursively the ``mind-stream'' of an LLM via a tree of self-generated follow-up questions.
Interestingly, by asking the LLM to generate a set of follow-up questions to its own answers creates (especially when the process recurses) a more focused ``stream of thoughts'', possibly as an emergent property of its ``in-context learning'' abilities.

\FIG{consneg}{DeepQA with ``{\em How constructive negation works in logic and constraint programming?}'' }{0.20}{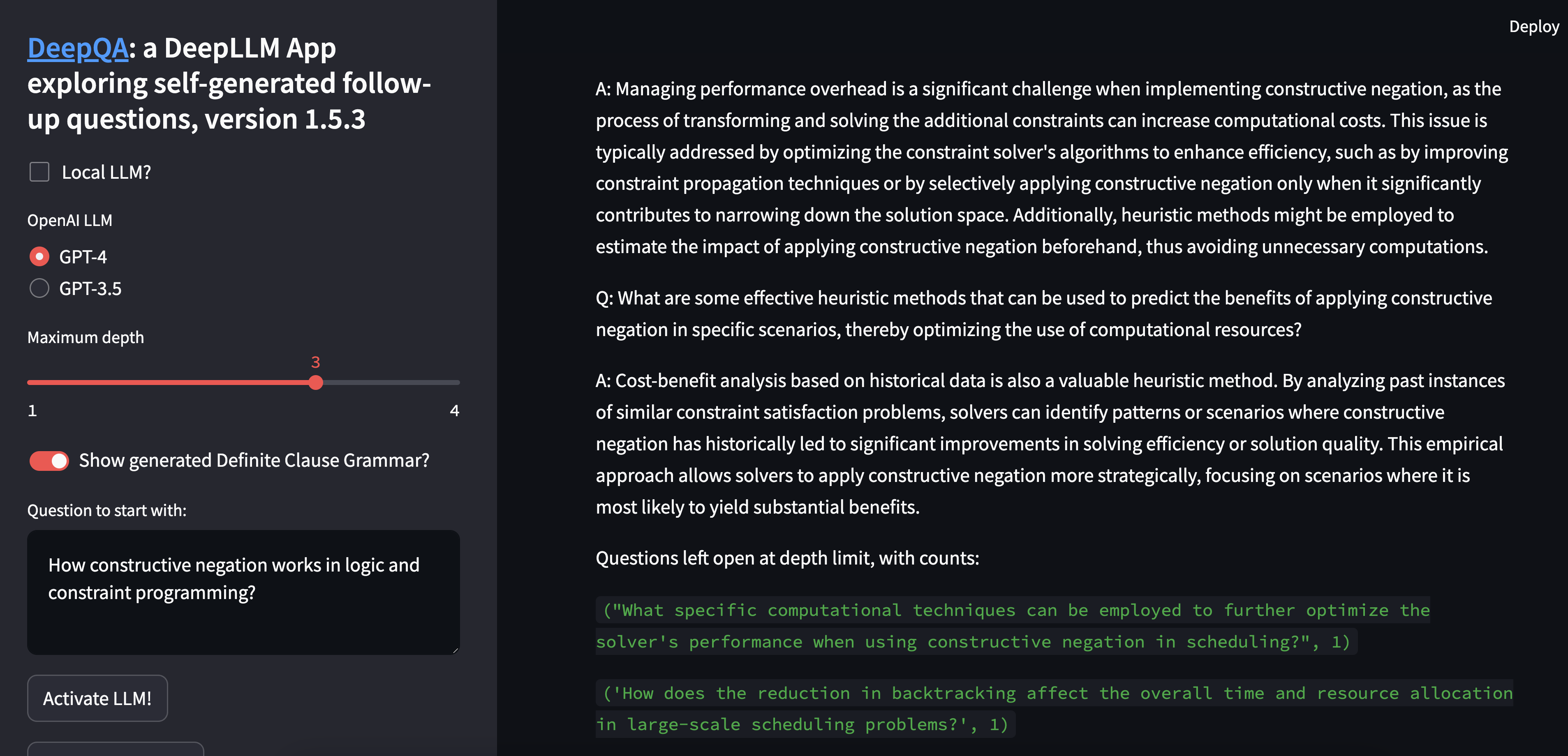} 

After started from an initiator question on a topic of the user's choice, the app explores its tree of {\em follow-up questions} up to a given depth. As output, it generates a Definite Clause Grammar that can be imported as part of a Prolog program. The DCG, in generation mode, will replicate symbolically the equivalent of the ``stream of thoughts'' extracted from the LLM interaction, with possible uses of the encapsulated knowledge in Logic Programming applications.

The synthesized grammar is designed to generate a finite language (by carefully detecting follow-up questions that would induce loops). We also ensure that paths in the question-answer tree are free of repeated answers, which get collected as well, together with questions left open as a result of reaching the user-set depth limit.

\BX
Definite Clause Grammar generated by initiator question {\em `How constructive negation works in logic and constraint programming?'}:

\begin{code}

q0-->q0_,a0_,q1.
q0-->q0_,a1_,q2.
q1-->q1_,a3_,q4.
...
q12-->q12_,a38_.
\end{code}
For instance, the first rule rewrites the initiator {\tt q0} into:
\BI
\I the terminal {\tt q0\_} that will produce the actual text of the question
\I the terminal {\tt a0\_} that will produce the actual text of the answer to {\tt q0\_}
\I the non-terminal {\tt q1} continuing the generation process with one of the follow-up questions generated by the LLM
\EI
\begin{code}

q0_-->['Q: How constructive negation works in logic and constraint programming?'].
q1_-->['Q: Can you provide an example of how constructive negation might refine 
   the solution space in a practical constraint programming problem?'].
q2_-->['Q: How does constructive negation differ from classical negation in terms 
   of computational efficiency and outcome in constraint satisfaction problems?'].
   ...
\end{code}

\begin{code}

a0_-->['A: Constructive negation in logic and constraint programming is a method 
    used to handle negation in a way that allows for the derivation of new 
    constraints from negative information. Instead of simply rejecting solutions 
    that do not satisfy a certain condition, constructive negation works by 
    deducing what must be true if a given condition is false. This is particularly
    useful in constraint programming where constraints define what is possible
    rather than what is not. By applying constructive negation, the system can
    infer additional constraints that must be met for the negation to hold,
    effectively refining the solution space.'].
    ...
\end{code}

When reaching the user-specified recursion depth, the unanswered follow-up questions are collected as ``open questions'' in the predicate {\tt opens/2} with the second argument indicating the number of times (over all branches of the tree) the question has been generated. In the case of an LLM with a very large parametric memory (e.g., GPT4, Claude 3 or Gemini) values above 1 are unlikely, while with smaller LLMs (e.g., Vicuna) repeated follow-up questions can happen more often.
\begin{code}

opens('What specific computational techniques can be employed to further optimize 
     the solver_s performance when using constructive negation in scheduling?',1).
opens('How does the reduction in backtracking affect the overall time and 
              resource allocation in large-scale scheduling problems?',1).
...
\end{code}
Starting the DCG in generation mode from its {\tt q0} initiator goal is achieved as follows:
\begin{code}
go:-q0(Xs,[]),nl,member(X,Xs),write(X),nl,nl,fail.
\end{code}

\EX
One can also use DeepQA to quickly assess the strength of an LLM before committing to it.
For instance, when used with a much weaker than GPT4 local LLM (enabled with Vicuna 7B by default) one will see shorter, more out of focus results, with a lot of repeated questions and answers collected by DeepQA in corresponding bins.

The full Prolog code discussed in thus example is available 
online\footnote{\url{https://github.com/ptarau/output_samples/tree/main/deepqa}}
as well the DeepQA app\footnote{\url{https://deep-auto-quests.streamlit.app/}}.

\section{Computing minimal models of LLM-generated logic programs}\label{torch}

\subsection{Minimal model computation with a GPU-friendly Torch-based Linear Algebra Algorithm}

At deeper recursion levels, the generated logic programs,  providing a symbolic representation of an LLM's parameter memory can quickly 
reach millions of clauses, ready to reason with.

To take advantage of the significant acceleration provided by GPUs we have implemented
a {\tt torch}-based linear algebraic minimal model computation 
algorithm\footnote{\url{https://github.com/ptarau/recursors/blob/main/tenslogic/proptens.py}} along the lines of \cite{linalg_lp}.

The implementation is centered around
\begin{code}
def tp(M, v):
    """
    one step fixpoint operator
    """
    r = M @ v
    return (r >= 1.0).to(torch.float32)
\end{code}
that advances one step of the fixpoint computation with a matrix multiplication ``\verb~@~'' and
\begin{code}
def tp_n(M, v0):
    """
    iterated fixpoint operator
    """
    oldv = v0
    n = M.shape[0]
    for i in range(n):
        newv = tp(M, oldv)
        if torch.allclose(newv, oldv):
            return newv
        oldv = newv
\end{code}
that proceeds until a fixpoint is detected using {\tt torch.allclose}.

The program contains readers of Horn clause programs represented in as {\tt .json} files. It can handle medium size programs (a few thousand clauses), as despite the GPU acceleration, complexity is still dominated by $O(N^3)$ matrix products.

We will show here a small test program running the minimal model computation. After defining:
\begin{code}
top = "true"
bot = "false"

vs = (p, q, r, s) = "pqrs"
\end{code}
We represent the program as pair made of the head of the clause and the list of atoms in its body:
\begin{code}
prog = [
        (p, [q]),
        (p, [r]),
        (q, [r, s]),
        (r, [top]),
        (bot, [q])
]
\end{code}
We can then compute the model with:
\begin{codex}
>>> print(compute_model(prog))
['p', 'r']
\end{codex}

Future work using torch sparse 
tensors\footnote{\url{https://pytorch.org/docs/stable/sparse.html}}, to ensure scalability for very large generated programs is planned along the lines of \cite{sparse2022}.

\subsection{Fixpoint-based minimal model computation}

It is not unusual to have loops in the propositional Horn Clause program connecting the LLM generated items by our recursors and refiners that would create problems with Prolog's depth-first execution model.
As using a SAT-solver would be an overkill in this case, given that Horn Clause and Dual Horn clause formula satisfiability is known to be polynomial,
we have implemented a simple low-polynomial complexity  \cite{dowling}
propositional satisfiability checker and model builder\footnote{\url{https://github.com/ptarau/recursors/blob/main/deepllm/horn_prover.py}}.

The model builder works by propagating truth from facts to rules until a fix point is reached. Given a Horn Clause $h:-b_1,b_2,...,b_n$, when all $b_i$ are known to be true (i.e., in the model), $h$ is also added to the model. If integrity constraints (Horn clauses of the form $false:-b_1,b_2,...,b_n$) have also been generated by the oracle agents monitoring our refiners, in the advent that all $b_1,b_2,...,b_n$ end up in the model,  $b_1,b_2,...,b_n$ implying $false$ signals a contradiction and thus unsatisfiability of the Horn formula associated to the  generated program.  However as the items generated by our recursive process are not necessarily expressing logically connected facts (e.g., they might be just semantic similarity driven associations), turning on or off this draconian discarding of the model is left as an option for the application developer.
Also, the application developer can chose to stop as soon as  a proof of the original goal emerges, in a way similar to goal-driven ASP-solvers like \cite{casp}, irrespectively to unrelated contradictions elsewhere in the program.

\section{Generating relation triplets for knowledge graphs}\label{rels}

Our DeepLLM app offers an option to generate from the minimal model of the program a relation graph (see {\bf Fig. \ref{relgraph}}) consisting of implication links (marked with ``{\tt :}'') to which it adds generalization links (marked with ``{\tt is}'').

\FIG{relgraph}{Relation graph for ``{\em tailgate when driving}'' }{0.48}{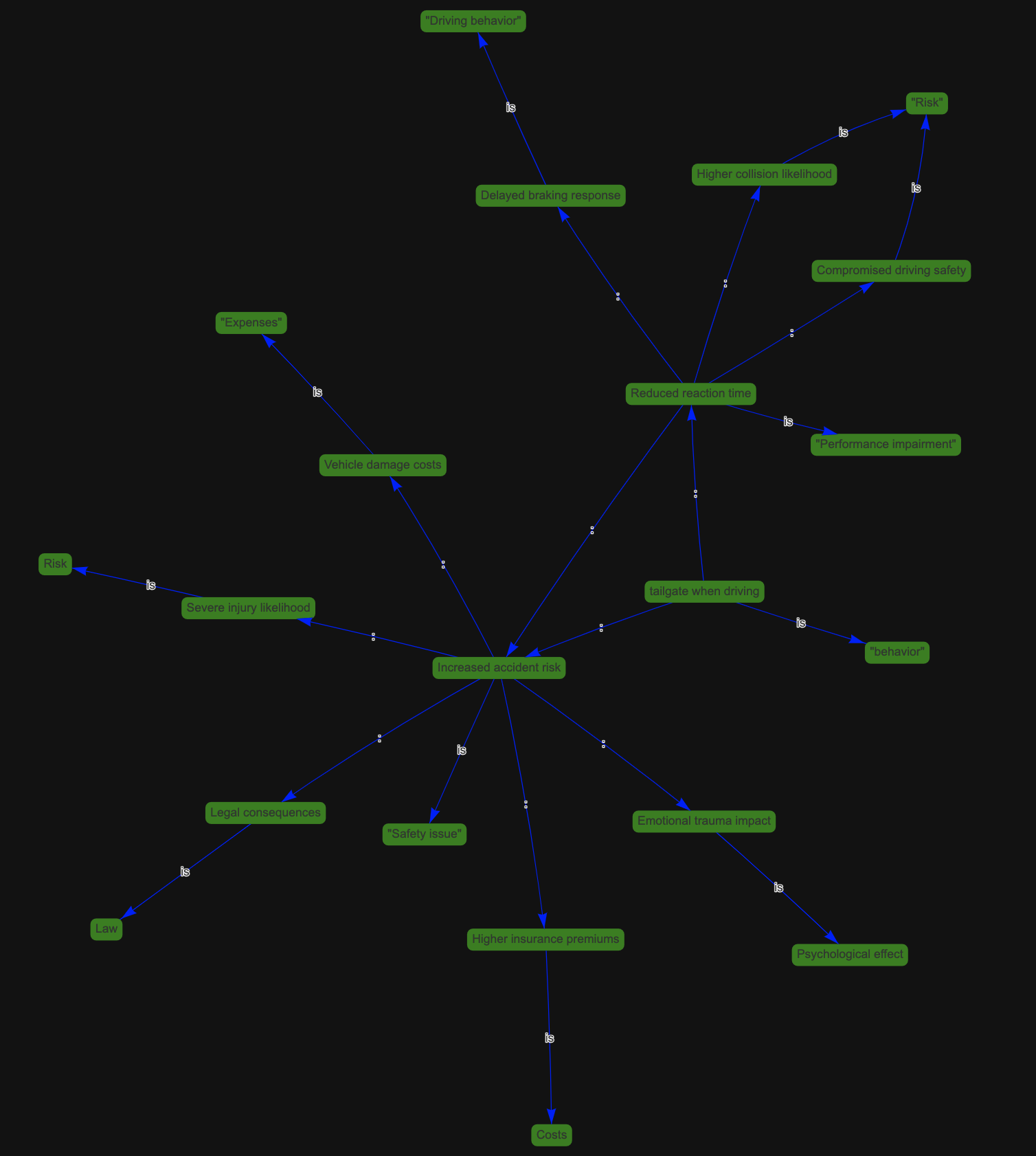} 

Implication links are extracted directly from the logic program while generalization links, serving as additional explanations, are generated by the LLM via an additional request.

Several other types of relation graphs can be generated depending on the planned reasoning method.

One of them is extraction of  \verb~<~subject, verb, object\verb~>~ (SVO) triplets obtained by prompting the LLM to split a complex sentence in simpler ones and extract from each simple sentence an SVO triplet.

Another is a hybrid method, combining relations extracted by using dependency grammars \cite{tplp21}, embeddings-based similarity relations, Wordnet-based and LLM-generated hypernyms and meronyms.

\section{Reasoning with soft unification on noisy facts}\label{soft}

The minimal models of LLM-generated Horn clause programs encapsulate
facts and their consequences elicited from DeepLLM's initiator queries
in the form of natural language sentences.
When writing a logic program that performs symbolic reasoning
relying on a ground fact database of such sentences, an interesting
form of {\em abductive reasoning} emerges.
When hitting an undefined ground sentence, intended as a query to match
database facts we can rely on vector embedding of the sentences and proximity
of the query and the facts in the vector space as a ``good enough'' match,
provided that the semantic distance between them is below a given threshold.
We will next describe a proof of concept of this strategy that we illustrate
on a small quotation dataset consisting of a few
sentences\footnote{\url{https://github.com/ptarau/natlog/blob/main/docs/quotes.txt}}.

We have implemented  {\tt Softlog}\footnote{\url{https://github.com/ptarau/natlog/tree/main/softlog}} as
an extension to the {\tt Natlog} system \cite{iclp21,iclp23natlog}, a Python-based 
Prolog dialect with a simpler syntax and a lightweight Python interface.
It works simply by overloading {\tt Natlog}'s built-in ground fact database's unification method with a form of {\em soft-unification} \cite{soft_unif,ltn22,deepsoftlog23}, implemented as follows:

\begin{code}
     def unify_with_fact(self, goal, trail):
     
        # q = query goal to be matched
        # k = number of knns to be returned
        # d = minimum knn distance
        # v = variable to be unified with the matches
        
        q, k, d, v = goal
        d = float(d) / 100
        _, answers = self.emb.knn_query(q, k)
        for sent, dist in answers:
            if dist <= d:
                self.abduced_clauses[(q, sent)] = dist
                yield unify(v, sent, trail)
\end{code}

The following {\tt Natlog} script is then used to query a small set of sentences
serving as {\tt Softlog}'s ground database. Note that the ``\verb|~|'' symbol is Natlog's convention for marking calls to a ground (soft-)database.
\begin{code}
knn 3.
threshold 70.

quest  Quest Answer:
  knn K, 
  threshold D,
  ~ Quest K D Answer.
\end{code}
We implement soft unification queries  as K closest neighbors (KNN) computations against embeddings in our  {\tt sentence\_store}\footnote{\url{https://github.com/ptarau/sentence_store}}. We use Sentence Transformers \cite{sbert} to compute embeddings and store them locally in an efficient and scalable vector database.
As usual in {\tt Natlog}, the Python iterator returning multiple KNN matches is mapped to Prolog's backtracking with multiple answers returned as alternative bindings to a result variable.
 
\begin{codex}
?- quest 'What happens if you do not know where you go' X?
ANSWER: {'X': 'If you don t know where you are going 
               you will end up somewhere else said Yogi Berra.'}
ANSWER: {'X': 'If you don t know where you are going
               any road will get you there said Lewis Carroll.'}
\end{codex}

When a query {\tt (Q A)} binds {\tt A} to an answer extracted from the vector store,
a binary clause {\tt (Q :- A)}  and its supporting fact {\tt (A :- true)} are inserted into the   dictionary of {\em abduced clauses}. If we add them to the program that triggered the generation of the clauses, we obtain a self-contained standard logic program that returns exactly the same answers as its {\tt Softlog} counterpart.
Alternatively, the computed distances  can be normalized as probabilities,  to annotate clauses used in a Probabilistic Logic Programming language like Problog \cite{de2007problog}.
\begin{codex}
ABDUCED CLAUSES:

'What happens if you do not know where you go' : 
    'If you don t know where you are going you will end up somewhere else 
    said Yogi Berra'. % distance=0.5874345302581787
'What happens if you do not know where you go' : 
    'If you don t know where you are going any road will get you there 
    said Lewis Carroll'. % distance=0.6724047660827637
\end{codex}
Note that by contrast to the usual exact unification based answers, {\tt Softlog} works quite well when the query is {\em close enough} to a matching entry in the sentence store, a reasonable assumption when the facts have been generated from multiple LLM runs and several ground truth resources.
\begin{codex}
?- quest 'What did Wilde say about temptation' X?
ANSWER: {'X': 'I can resist anything except temptation said Oscar Wilde.'}

?- quest 'What did Alice say about following advice' X?
ANSWER: {'X': 'I give myself very good advice but I very seldom follow it 
               said Lewis Carroll.'}
\end{codex}
Given the nature of semantic search, surname is enough to find Oscar Wilde and as {\tt Alice} associates with the author Lewis Carroll, soft unification will  fetch it from the sentence store.

\section{Related Work}\label{rel}

By contrast to ``neuro-symbolic'' AI \cite{neurosym}, where the neural architecture is closely intermixed with symbolic steps, in our approach the neural processing is encapsulated in the LLMs and accessed via a declarative, high-level API. This reduces the semantic gap between the neural and symbolic sides as their communication happens at a much higher, fully automated and directly explainable level.

Our recursive descent algorithm shares the goal of extracting more accurate information from the LLM interaction with work on ``Chain of Thought'' prompting of LLMs \cite{chainofthought,ling2023deductive} and with step by step \cite{lightman2023lets} refinement of the dialog threads. 
Our approach shares with tools like LangChain \cite{langchain} the idea of piping together multiple instances of LLMs, computational units, prompt templates and custom agents, except that we fully automate the process without the need to manually stitch together the components.

We have not found any references to the use of Dual Horn clauses in logic programming but it is a well known result \cite{schaefer78}) that their complexity in the propositional case is polynomial, similarly to their of Horn clause counterparts. This fact makes them also good generation targets for LLM-extracted knowledge processing. 

We have not found anything similar to generating question-answer-follow-up question chains, although it is common practice for chatbots to  suggest (a choice between) follow-up questions\footnote{ including the author's own \url{https://auto-quest.streamlit.app/}}.

Our torch-based model-computation algorithm follows closely the matrix-computation logic of \cite{linalg_lp}, our contribution being its succinct and efficient GPU-friendly implementation.

Interest in several forms of soft-unification has been active  \cite{soft_unif,ltn22,deepsoftlog23} as differentiable substitute of symbolic unification in neuro-symbolic systems.
By contrast, our focus in this paper is flexible information retrieval of LLM-generated natural language content, for which high quality embeddings were available either from LLM APIs or local resources like the torch-based sentence-transformers \cite{sbert}.

\section{Conclusion}\label{conc}

It is now undeniable that Generative AI is a major disruptor not just of industrial fields ranging from search engines, automation of software development and robotics to medical and legal advisory systems,
but also a disruptor of research fields, including symbolic AI as we know it and machine Learning itself. In particular, results produced by dominant ML or NLP techniques as well as work on integration of neural and symbolic systems have become replaceable
by much simpler applications centered around LLM queries and RAG systems.
In fact, by concentrating the knowledge encapsulated in its parametric memory into a single declarative interface, Generative AI can replace complex, labor-intensive  software functionality with a simple LLM API call or a question in one's favorite natural language.

This motivates our effort to ``join the disruption'' and explore several new
ways to elicit the knowledge encapsulated in the LLMs' parametric memory
as logic programs, together with an investigation of their optimal
inference execution methods. We have not just exposed as logic programs 
the several kinds of
knowledge snippets extracted by recursive automation LLM dialog threads , but we have also devised efficient inference execution mechanisms for them.

We hope that this effort has revealed some natural synergies between  Generative AI systems and logic programming tools, ready to fill  gaps like the lack of rigorous reasoning abilities of the LLMs, their lack of alignment to the user's intents and their known deficiencies on factuality.

\bibliographystyle{eptcs}

\bibliography{tarau,ml,proglang,theory}

\end{document}